\documentclass[12pt,reqno]{article}
\usepackage[english]{babel}
\usepackage{amsmath,dsfont, amsthm, amssymb}
\usepackage{amsfonts}
\usepackage{graphicx}
\usepackage{enumerate}
\usepackage[usenames,dvipsnames]{color}
\usepackage{subfigure}
\usepackage{subcaption}
\usepackage[right,pagewise,displaymath, mathlines]{lineno}
\usepackage{epstopdf}
\usepackage{color}
\usepackage{multirow}
\usepackage[table,xcdraw]{xcolor}
\usepackage[title]{appendix}

\usepackage{authblk}
\usepackage[round]{natbib}
\usepackage{float}

\usepackage{enumitem}
\usepackage{soul}

\usepackage{tikz}
\usepackage{schemabloc}

\usepackage[bottom]{footmisc}

\usepackage[colorlinks = true,
            linkcolor = blue,
            urlcolor = blue,
            citecolor = blue,
            breaklinks]{hyperref}

\usepackage{geometry}
\numberwithin{equation}{section}
\numberwithin{figure}{section}
\numberwithin{table}{section}

\theoremstyle{definition}

\numberwithin{equation}{section}

\definecolor{darkread}{rgb}{0.7, 0, 0}
\definecolor{darkbrown}{rgb}{0.55, 0.2, 0.15}
\definecolor{darkblue}{rgb}{0.1,0.1,0.6}
\definecolor{darkgreen}{rgb}{0.1,0.5,0.2}

\usepackage{setspace}
\usepackage[bottom]{footmisc}

\usepackage{tkz-graph}
\usepackage{pgfplots}

\title{\vspace{-15mm}Distributional sentiment modeling and anomaly detection for consumer complaint assessment
\thefootnote\relax\footnotetext{
This research was initiated by Ri\v cardas Zitikis (1963-2025) in early 2025. This article is also dedicated to his memory.}
}

\author[,1]{\normalsize Peiheng Gao\thanks{Corresponding author; e-mail: \href{mailto:pgao47@uwo.ca}{pgao47@uwo.ca}}}

\author[2,3]{Chen Yang}

\author[4]{Shimin Zhang}

\affil[1]{\normalsize School of Mathematical and Statistical Sciences, Western University, London, ON~N6A~3K7, Canada}

\affil[2]{\normalsize Department of Population Health Science and Policy, Icahn School of Medicine at Mount Sinai, New York, New York~10029, USA}

\affil[3]{\normalsize Institute for Health Care Delivery Science, Icahn School of Medicine at Mount Sinai, New York, New York~10029, USA}

\affil[4]{\normalsize Asian Institute of Digital Finance, National University of Singapore, Singapore, 119077}

\date{}
\begin{document}

\maketitle

\begin{abstract}
Sentiment analysis is a common tool for converting unstructured text into quantitative signals in finance and risk management. Yet most applications reduce the output to a discrete polarity label or a single predictive feature, overlooking the distributional structure of sentiment intensity in consumer complaint narratives. In this paper we treat negative sentiment in consumer complaints as a bounded continuous variable and study its full distribution rather than a single label. We score each narrative with a transformer classifier, model the scores with Beta distributions, and compare the fitted distributions of meritorious and non-meritorious complaints through the Kullback–Leibler divergence and the squared Hellinger distance. The fitted distributions are then linked with dollar amounts and company response outcomes to construct anomaly diagnostics that flag complaints whose textual severity is inconsistent with the recorded relief. We find that the two groups have strongly overlapping distributions, so negative sentiment intensity is not a sharp classifier of outcomes on its own; combined with monetary and categorical attributes, it isolates unusually severe complaints for operational risk monitoring. Treating sentiment analysis as continuous distributional measurement, this study links sentiment extraction, bounded response modeling, and anomaly detection in a unified framework for consumer complaint assessment.
\end{abstract}

\medskip



\section{Introduction}
The volume of digitally recorded text keeps growing, and much of it carries information that structured data alone does not capture. Regulatory filings, consumer complaint narratives, financial news, and social media posts describe events, judgments, and experiences in natural language. A basic task is therefore to turn unstructured text into numerical signals for statistical modeling. Sentiment analysis (SA) serves these type of tasks by assigning a numerical score to text, and in many applications this score records the affective orientation of the text on a bounded scale \citep{Gao2025b}.

Consumer complaint narratives offer a useful setting for this task. Such narratives record dissatisfaction, dispute frictions, service failures, and potential operational weaknesses, and they use heterogeneous language, including informal expressions, terms specific to particular products, and detailed accounts of financial harm. As a consequence, complaint narratives are not only inputs for classification, they are also measurement objects for the intensity and distribution of consumer dissatisfaction.

Much of the applied SA literature reduces sentiment scores to discrete polarity labels before further analysis. This step supports document classification, yet it keeps only part of the information in the sentiment output. The present study instead treats the negative sentiment score of each complaint narrative as a continuous variable on $(0,1)$. This view preserves the numerical precision of the score and allows the distribution of sentiment intensity to be studied directly through the mean, variance, shape, and tail behavior of the scores, and across complaint groups, products, or resolution outcomes.

Because SA scores are bounded, the Beta distribution provides a natural model for sentiment intensity. Its two shape parameters, $\alpha$ and $\beta$, represent symmetric, skewed, concentrated, and dispersed patterns on $(0,1)$, so a fitted Beta distribution gives a compact summary of the sentiment profile of a complaint group. Fitted distributions can then be compared through distributional distances (e.g., the Kullback--Leibler (KL) divergence and the squared Hellinger distance). The same fitted distributions support record level anomaly diagnostics, since a complaint may fall in the tail of its group distribution or may carry a sentiment score that is inconsistent with its observed outcome. These cases matter for operational risk monitoring because they reveal patterns that standard classification metrics may not capture.

The present study applies this distributional view to consumer complaint narratives from the Consumer Financial Protection Bureau (CFPB). Following the outcome construction in \citet{GYSZ2026}, complaints closed with monetary or non-monetary relief are labeled meritorious, and complaints closed with explanation are labeled non-meritorious. The central question is whether meritorious and non-meritorious complaints differ in their full negative sentiment distributions, and not only in average sentiment or predicted class labels.

\subsection{Related studies}
\subsubsection{Sentiment analysis and complaint narratives}
SA is widely used to turn text into quantitative signals, with applications in customer reviews, social media monitoring, financial text analysis, and service quality assessment. Complaint data suit SA well because complaints record dissatisfaction, service failures, dispute frictions, and operational weaknesses. Recent work on CFPB narratives shows that textual signals support outcome prediction and risk monitoring in consumer finance, including the prediction of monetary relief outcomes \citep{WZZ2026}. Related work also shows that sentiment often attaches to specific aspects of a narrative, such as fees, fraud disputes, credit reporting, customer service, or account servicing, rather than to the whole document \citep{DZD2024,ZHZ2024,WMZ2024}. 

Recent studies further show that SA has moved from coarse polarity classification toward modeling that is specific to a domain and aware of context. For example, large volumes of YouTube news comments have been analyzed through sentiment, emotion, hate speech, and topic modeling to characterize public discourse \citep{villaperez2026public}. In financial settings, sentiment from related firms can improve stock price forecasting, suggesting that sentiment signals may propagate through connected entities rather than remaining isolated at the document or firm level \citep{chen2026dual}. Broader NLP surveys also show that models adapted to specific domains and large language models improve text based assessment tasks, although they raise practical concerns about cost, scalability, and deployment \citep{cao2026mentalhealth}. SA thus offers a natural route from complaint text to measurable signals, while the complaint setting calls for finer detail than a coarse positive, neutral, and negative classification.

\subsubsection{Operational risk and financial text analytics}
Operational risk research increasingly moves from backward looking loss recording toward earlier detection of weak signals in unstructured text, since many risk events first appear in narratives, incident reports, and complaints before they enter coded loss databases. \citet{Ji2023OpRiskKG} combine text mining, analytic hierarchy process assessment, and knowledge graphs to analyze operational risk records. Building on this line, \citet{Piacenza2026} propose an integrated workflow that combines internal event narratives with external web content, including near real time monitoring of posts on X, to flag emerging and novel operational risks. In consumer finance, \citet{Gao2024} detect systematic anomalies among CFPB complaint narratives, which motivates the present focus on complaint text as a risk bearing signal, and \citet{CCKP2024} show that combining textual, vocal, and market information improves financial risk prediction. 

Machine learning has also been used in adjacent operational settings where risk is observed through large volumes of business records. \citet{barros2026supply} propose a framework based on big data for predicting supplier delay risk and show that predictive models should be evaluated not only by error metrics but also by their downstream operational costs. This reinforces the view that signals from text and data are useful only when they support interpretable risk monitoring and decision making. Broader surveys of financial large language models report gains on financial text tasks alongside concerns about calibration, hallucination, and cost \citep{LSHS2024}. Together, this literature shows that unstructured text can reveal emerging risks and institution level weaknesses that structured variables miss, yet it mostly targets point predictions or case level flags and rarely models the distribution of the sentiment scores themselves.

\subsubsection{Bounded sentiment scores, Beta models, and topic representations}
Once narratives become sentiment scores, the statistical form of those scores matters. Many sentiment models produce outputs that behave like probabilities, such as negative, neutral, and positive probabilities, and these values lie in the unit interval. The Beta distribution therefore suits sentiment intensity as a bounded continuous variable. Recent studies show that beta regression serves rates, proportions, and other responses restricted to $(0,1)$ \citep{FGM2024}, with extensions that handle boundary values at 0 and 1 \citep{KZ2024} and scalable, robust models for continuous proportional data \citep{LDOD2025}. Models based on the Beta distribution also appear in Bayesian optimization over bounded domains \citep{NZBT2025}, and biomedical image learning with differentiable Beta noise masks \citep{YY2024}. Bounded sentiment modeling also relates to topic representations. Latent Dirichlet allocation (LDA) and related topic models decompose a corpus into latent themes, and recent work shows that topic modeling still helps monitor customer discussions in financial services \citep{OMHGB2024} and that latent category distributions capture overlapping and detailed sentiment structure \citep{ZYASZLDJ2024}. Both the Beta and Dirichlet distributions describe quantities on bounded domains: the Dirichlet distribution models probability vectors on a simplex, while the Beta distribution models a single proportion or probability. For complaint narratives, LDA describes latent issue proportions, while the Beta distribution describes continuous sentiment intensity on $(0,1)$, yet the two are seldom combined in a single distributional treatment of complaints.

\subsubsection{Distributional distances and divergences in sentiment analysis}
The distributional view leads to comparisons of fitted sentiment distributions across groups. The KL divergence measures directional information loss between two distributions, while the squared Hellinger distance gives a symmetric and bounded measure of separation. Recent work confirms the practical value of both. KL divergence serves as an anomaly filter that needs no model for noisy sensor series \citep{ZGS2024}, and as a tool for approximating or diversifying the output distributions of language models \citep{WOLSG2024}. The squared Hellinger distance supports upper confidence bounds for stochastic bandits and for cold start problems in recommender systems \citep{YWM2024}, the accuracy of density estimation \citep{KCR2024}, and robust estimators based on the minimum Hellinger distance for complex survey designs \citep{KV2025}. For sentiment analysis, these measures shift the focus from pointwise polarity classification to comparison of group distributions. The question is not only whether a narrative is negative, but whether the full distribution of negative sentiment intensity differs between meritorious and non-meritorious complaints. Fitted Beta distributions summarize each group, KL divergence measures directional information loss between them, and the squared Hellinger distance measures their symmetric separation, yet such distributional comparison remains uncommon in complaint analytics.

\subsubsection{Anomaly detection in sentiment analysis}
Anomaly detection extends SA when the goal is to flag unusual narratives or cases whose sentiment profile differs from the observed outcome. Recent surveys link anomaly detection and the identification of inputs outside the training distribution to large language models, which supply semantic representations and contextual reasoning for unusual textual patterns \citep{XD2024}. A systematic review of forecasting and anomaly detection based on large language models further shows that textual signals can help reveal abnormal behavior, with continuing attention to generalizability, explainability, hallucination, and computational cost \citep{SJJQXMWJXL2024}. 

Recent anomaly detection studies also stress the value of combining several signals or models when abnormal cases are rare and labels are scarce. For attributed graphs, \citet{khan2026context} detect anomalies by jointly modeling node attributes, neighborhood structure, and information at the community level, so that unusual nodes are judged against several complementary views rather than a single feature. For public sector auditing, \citet{schiavon2026parliamentary} apply an unsupervised ensemble to flag suspicious Brazilian parliamentary expenses from unlabeled reimbursement records, and show that a conservative consensus across detectors raises precision for later human review. Both settings share the goal of the present paper: when labels are limited, the task is less to classify every case than to rank unusual cases for closer inspection. This motivates diagnostics that read each record against a fitted distribution and surface the records that depart from it, rather than relying on a single hard label.

In sentiment applications, anomaly detection can operate at the record level: a complaint may carry an unusually high or low negative sentiment score, or its observed outcome may be inconsistent with the fitted sentiment distribution under which the score is most plausible. These cases motivate distributional diagnostics based on Beta tail probabilities, KL divergence, and the squared Hellinger distance, which remain underused for complaint assessment.
\subsection{The scope}

The present study develops a distributional approach to negative sentiment intensity in consumer complaint narratives, together with record level diagnostics for anomalies. Most SA studies treat sentiment as a discrete label or a single predictive feature, and recent complaint studies center on outcome prediction. Beta family models, topic representations, and distributional distances together allow bounded continuous scores to be modeled and full distributions to be compared. The present study builds on this literature by treating the negative sentiment score of each complaint narrative as a continuous variable on $(0,1)$, and by examining whether meritorious and non-meritorious complaints differ in average sentiment, distributional shape, dispersion, and tail behavior. The scope covers CFPB narratives, group Beta fitting, distributional comparison through the KL divergence and the squared Hellinger distance, and sentiment diagnostics for outcome inconsistency, tail behavior, and high negative sentiment cases without relief. A full topic model of complaint issues is outside the scope of the present paper.

\subsection{The objective}
Within this scope, the objective of the present study is to build a distributional framework for continuous sentiment intensity in consumer complaint narratives and to use it for record level anomaly diagnostics. The analysis proceeds in four steps.
\begin{enumerate}
    \item Continuous negative sentiment scores are constructed for the cleaned narratives, and a boundary correction is applied so that the scores suit Beta family modeling.

    \item A Beta distribution is fitted to the full cleaned sample as a baseline and, separately, to the sentiment scores of the meritorious and non-meritorious groups, which yields a compact summary of the sentiment profile of each.

    \item The separation between the two fitted distributions is measured with the KL divergence and the squared Hellinger distance, which tests whether negative sentiment intensity carries group information beyond pointwise classification.

    \item Diagnostics based on the fitted distributions are used to identify outcome sentiment inconsistency, tail behavior, and high negative sentiment cases without relief.
\end{enumerate}

\subsection{Contributions}
The present study contributes to work on consumer complaint analytics, financial text mining, and operational risk assessment in five ways.
\begin{enumerate}
    \item The present paper reframes complaint sentiment as a distributional object. Negative sentiment intensity is treated as a continuous random variable on $(0,1)$ rather than as a polarity label, so that complaint sentiment can be studied through its whole distribution.
    \item The present paper contributes one of the first applications of Beta family distributional modeling to compare and characterize meritorious and non-meritorious complaints. The fitted models give interpretable representations of bounded support, asymmetry, and dispersion in the observed SA scores.
    \item An evaluation layer based on the KL divergence and the squared Hellinger distance is introduced to compare the two fitted distributions. This layer complements classification metrics with a measure of distributional separation.
    \item Sentiment diagnostics for anomalies connect the analysis to operational risk monitoring. Outcome sentiment inconsistency, tail observations, and high negative sentiment cases without relief are read as screening signals rather than as classification errors.
    \item The procedure yields a transparent diagnostic layer for regulators, financial institutions, and consumer protection agencies. It shows whether sentiment distribution features aid the interpretation of meritorious and non-meritorious complaints and supports downstream complaint classification and risk assessment.
\end{enumerate}
Overall, the present study moves the role of SA in complaint assessment from pointwise classification toward distributional reliability analysis. This move suits complaint narratives, whose signals depend on context and appear both in individual sentiment scores and in the distribution of scores across outcomes. The rest of the present paper is organized as follows. Section~\ref{Data} describes the CFPB complaint data and the cleaning procedure. Section~\ref{Method} presents the distributional sentiment methodology, including sentiment score construction, Beta fitting, divergence measures, and anomaly diagnostics. Section~\ref{resultsC} reports the empirical distributional findings. Section~\ref{conclusioned} concludes the present paper.

\section{Data}
\label{Data}
The present study follows earlier CFPB studies for the data structure, dollar amount extraction, and outcome label construction \citep{Gao2025}. Each record contains structured fields, the company response, product and issue categories, and, when the consumer allows it, a written complaint narrative \citep{CFPB}.
The sample covers complaints filed between 1 September 2023 and 30 June 2026. The start date is chosen mainly because the CFPB revised the complaint form on 24 August 2023, changing the product, subproduct, issue, and subissue options \citep{CFPB}. Starting on 1 September 2023 keeps the product and issue labels consistent after the form change. The date also falls after the end of the main COVID-19 emergency period, since the U.S. public health emergency ended on 11 May 2023 and the World Health Organization ended its emergency declaration on 5 May 2023 \citep{HHS2023PHE,WHO2023COVID}.

\subsection{Preprocessing and visualization}
The date filter leaves $2{,}280{,}111$ records. which around $72.92\%$ dataset fall under the product category ``Credit reporting or other personal consumer reports''. These reports affect access to credit, housing, and employment screening, so the present study keeps only this category.
The next step needs a dollar amount and an outcome label. Let $A_i$ be the dollar amount taken from narrative $i$. When a narrative lists several amounts, $A_i$ is their average. A narrative stays in the sample when it meets three rules.
\begin{enumerate}
    \item The narrative contains at least one dollar amount.
    \item The average amount satisfies $1\leq A_i<10{,}000$.
    \item The company response marks the complaint as meritorious or non-meritorious.
\end{enumerate}
The upper cutoff drops averages of  $10{,}000$ dollars or more. Such large amounts often come from business accounts, institutional disputes, or unusual cases, not from a single consumer \citep{Gao2024}.

Table~\ref{data_cleaning_summary} 
\begin{table}[htbp]
\centering
\caption{Data cleaning steps and record counts.}
\label{data_cleaning_summary}
\begin{tabular}{lr}
\hline
Step & Records \\
\hline
After date restriction & $2{,}280{,}111$ \\
Credit reporting product category & $1{,}662{,}544$ \\
At least one detected dollar amount & $222{,}261$ \\
Average dollar amount in $1 \leq A_i<10{,}000$ & $181{,}316$ \\
\quad Meritorious & $85{,}501$ \\
\quad Non-meritorious & $95{,}815$ \\
\hline
Imbalance ratio & $1.12$ \\
\hline
\end{tabular}
\end{table}
lists the counts at each step. Among the credit reporting records, $222{,}261$ contain a dollar amount, and $181{,}316$ of these fall in the kept range. This last set is the cleaned sample $\mathcal{D}_{\mathrm{clean}}$, with $N_{\mathrm{clean}}=181{,}316$ narratives. Among them, $85{,}501$ are meritorious, that is, closed with monetary or non-monetary relief, and $95{,}815$ are non-meritorious, that is, closed with an explanation. The two groups are close in size, with an imbalance ratio  $1.12$. The baseline analysis therefore uses no oversampling approach. Oversampling, class weighting, and threshold adjustment serve as robustness checks.

\begin{figure}[htbp]
\centering
\includegraphics[width=\textwidth]{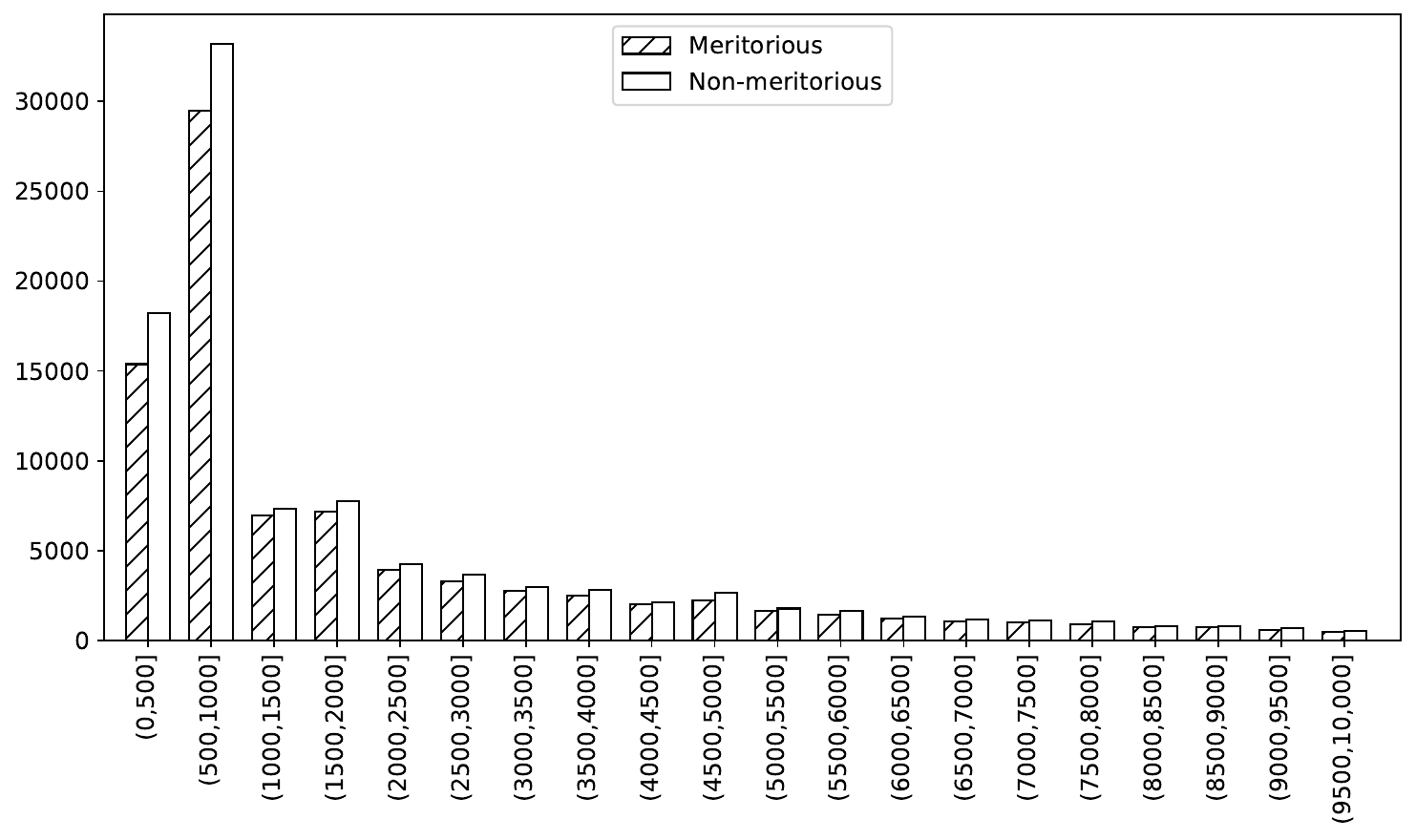}
\caption{Distribution of average dollar amounts by complaint outcome.}
\label{Fig1Result}
\end{figure}

Figure~\ref{Fig1Result}
shows how the average dollar amounts spread out within each group. Both groups cluster at low amounts, and the counts fall quickly as the amount grows, where the x-axis represent the average dollar amount and the y-axis represent the number of narratives under each different bins. The two groups share a similar right tail. Non-meritorious complaints are a little more common in some of the low bins. This shape fits the cleaned sample and supports the cutoff at $10{,}000$ dollars.

\section{Distributional sentiment modeling}
\label{Method}
The methodology converts cleaned complaint narratives into continuous negative sentiment signals and uses the extracted dollar amount \(A_i\) as monetary context, and the sentiment score measures textual intensity. These signals are developed through a five components research design. Each narrative is converted into a negative sentiment probability, corrected to lie in \((0,1)\), and modeled with Beta distributions for the full sample and outcome groups. The fitted group distributions are then compared using the KL divergence and the squared Hellinger distance. Finally, record level anomaly diagnostics are constructed from outcome sentiment inconsistency, tail probabilities, and high sentiment no relief cases. Algorithm~\ref{algorithm_distributional_sentiment}
\begingroup
\renewcommand{\tablename}{Algorithm}
\begin{table}[htbp]
\centering
\caption{Algorithmic summary of the distributional sentiment modeling.}
\label{algorithm_distributional_sentiment}
\begin{tabular}{l}
\hline
\textbf{Require} \\
1. Cleaned complaint narrative corpus \(\mathcal{C}_{\mathrm{clean}}\). \\
2. Complaint outcome labels \(Y_i \in \{0,1\}\), where \(Y_i=1\) denotes meritorious. \\
3. Extracted average dollar amounts \(A_i\) satisfying \(1 \leq A_i<10{,}000\). \\
\hline
\textbf{Compute distributional sentiment and anomaly diagnostics} \\
1. Apply a transformer based sentiment classifier to each narrative \(c_i\). \\
2. Extract the negative sentiment probability \(s_i=p_{i,\mathrm{neg}}\). \\
3. Apply the boundary correction to obtain \(s_i^* \in (0,1)\). \\
4. Fit Beta distributions to the full sample and to each outcome group. \\
5. Evaluate outcome group separation using KL divergence and squared Hellinger distance. \\
6. Compute record level anomaly diagnostics from outcome sentiment inconsistency, tail probabilities,  \\ and high sentiment no relief cases. \\
\hline
Output \\
1. Corrected negative sentiment scores \(\{s_i^*: i \in \mathcal{D}_{\mathrm{clean}}\}\). \\
2. Fitted Beta parameters for the full sample and outcome groups. \\
3. Distributional separation measures between complaint outcome groups. \\
4. Record level anomaly indicators and diagnostic rates. \\
\hline
\end{tabular}
\end{table}
\endgroup
summarizes the workflow, and the following subsections describe each component.

\subsection{Continuous sentiment score construction}
SA assigns a numerical score to each document in a corpus, thereby converting unstructured textual content into a quantitative signal suitable for statistical modeling. Let \(\mathcal{D}_{\mathrm{clean}}\) denote the cleaned sample defined in the Section~\ref{Data}, with \(N_{\mathrm{clean}}\) complaint narratives. For each observation \(i \in \mathcal{D}_{\mathrm{clean}}\), let \(c_i\) denote the complaint narrative, \(A_i\) the extracted dollar amount, and \(Y_i \in \{0,1\}\) the complaint outcome label, where \(Y_i=1\) denotes a meritorious complaint and \(Y_i=0\) denotes a non-meritorious complaint. The cleaned narrative corpus is written as
\begin{equation*}
    \mathcal{C}_{\mathrm{clean}}=
    \{c_i: i \in \mathcal{D}_{\mathrm{clean}}\},\qquad
    |\mathcal{C}_{\mathrm{clean}}| = N_{\mathrm{clean}}.
\end{equation*}

The present paper uses a transformer based sentiment classifier to construct the sentiment score. Transformer models such as BERT, RoBERTa, and FinBERT encode sentence context and return class probabilities through a softmax output layer \citep{Devlin2019,Liu2019,Araci2019}. This structure is aligned with the distributional framework of the present paper because the negative class probability can be used as a bounded continuous sentiment score. A financial or domain adapted transformer model is prioritized because complaint narratives contain financial products, credit reporting terminology, dispute descriptions, and regulatory language. Following this choice, the present paper adopts the FinBERT sentiment scoring used in \citet{GYSZ2026}.

For each narrative \(c_i \in \mathcal{C}_{\mathrm{clean}}\), the transformer classifier returns the probability vector
\begin{equation*}
    \mathbf{p}_i=
    \bigl(p_{i,\mathrm{neg}}, p_{i,\mathrm{neu}}, p_{i,\mathrm{pos}}\bigr)
    \in \Delta^2, \qquad 
    \Delta^2
    =
    \{\mathbf{p} \in \mathbb{R}^3:
    p_k \geq 0,\;
    \sum_{k} p_k = 1\}.
\end{equation*}
The three components denote the model assigned probabilities of negative, neutral, and positive sentiment, respectively.

Because CFPB complaint narratives describe adverse consumer experiences with financial institutions, negative sentiment intensity carries the most direct signal for complaint severity and operational risk. The present paper therefore defines the continuous sentiment score as the negative sentiment probability,
\begin{equation*}
    s_i=\mathrm{SA}(c_i)=p_{i,\mathrm{neg}},\qquad 0 \leq s_i \leq 1.
\end{equation*}
A higher value of \(s_i\) indicates stronger negative sentiment in complaint narrative \(c_i\). This construction retains the numerical information in the classifier output rather than reducing sentiment to a discrete polarity label.

The Beta distribution has open support on \((0,1)\), while softmax probabilities may take boundary values close to 0 or 1. To ensure compatibility with Beta modeling, the boundary correction of \citet{Smithson2006} is applied:
\begin{equation*}
    s_i^*=\frac{s_i(N_{\mathrm{clean}}-1)+0.5}{N_{\mathrm{clean}}}.
\end{equation*}
This transformation maps sentiment scores into \((0,1)\) while preserving the ordering of observations. The corrected scores \(\{s_i^*: i \in \mathcal{D}_{\mathrm{clean}}\}\) are used in the distributional analysis below.

\subsection{Beta models for continuous sentiment intensity}
The corrected negative sentiment score \(s_i^*\) is treated as a bounded continuous random variable. Since \(s_i^*\in(0,1)\), the Beta distribution provides a natural parametric model for sentiment intensity. For a generic sample subset \(\mathcal{G}\subseteq\mathcal{D}_{\mathrm{clean}}\), the corrected scores are modeled as
\begin{equation*}
    s_i^*\sim\mathrm{Beta}(\alpha_{\mathcal{G}},\beta_{\mathcal{G}}),\qquad
    i\in\mathcal{G}.
\end{equation*}
The corresponding probability density function is
\begin{equation}
    \label{eq:beta_pdf}
    f(s_i^* \mid \alpha,\beta)=\frac{(s_i^*)^{\alpha-1}(1-s_i^*)^{\beta-1}}
    {B(\alpha,\beta)},\qquad 0<s_i^*<1,
\end{equation}
where \(B(\alpha,\beta)=\Gamma(\alpha)\Gamma(\beta)/\Gamma(\alpha+\beta)\) is the Beta normalizing constant.

The parameters for any subset \(\mathcal{G}\) are estimated by maximum likelihood:
\begin{equation}
    \label{eq:beta_mle}
    (\hat{\alpha}_{\mathcal{G}},\hat{\beta}_{\mathcal{G}})=
    \arg\max_{\alpha>0,\beta>0}\sum_{i \in \mathcal{G}}
    \log f(s_i^* \mid \alpha,\beta).
\end{equation}
When \(\mathcal{G}=\mathcal{D}_{\mathrm{clean}}\), the fitted distribution \(\mathrm{Beta}(\hat{\alpha},\hat{\beta})\) gives a compact description of the overall negative sentiment landscape in the cleaned CFPB complaint corpus.

To examine whether sentiment intensity differs by complaint outcome, the cleaned corpus is also partitioned into meritorious and non-meritorious subsets:
\begin{equation*}
    \mathcal{D}_{1}=\{i \in \mathcal{D}_{\mathrm{clean}}: Y_i=1\},
    \qquad \mathcal{D}_{0}= \{i \in \mathcal{D}_{\mathrm{clean}}: Y_i=0\}.
\end{equation*}
The corresponding group sizes are \(N_1=|\mathcal{D}_{1}|\) and \(N_0=|\mathcal{D}_{0}|\), with \(N_1+N_0=N_{\mathrm{clean}}\). Applying \eqref{eq:beta_mle} to these two subsets gives the fitted group specific distributions
\begin{equation*}
    s_i^* \mid Y_i=1\sim
    \mathrm{Beta}(\hat{\alpha}_1,\hat{\beta}_1),
    \qquad s_i^* \mid Y_i=0
    \sim \mathrm{Beta}(\hat{\alpha}_0,\hat{\beta}_0).
\end{equation*}
The difference between \((\hat{\alpha}_1,\hat{\beta}_1)\) and \((\hat{\alpha}_0,\hat{\beta}_0)\) provides distributional evidence on whether meritorious and non-meritorious complaints differ in negative sentiment intensity, dispersion, and shape. These fitted group distributions are the inputs for the separation metrics and anomaly diagnostics that follow.

\subsection{Distributional separation metrics}
Given the fitted group specific distributions \(\mathrm{Beta}(\hat{\alpha}_1,\hat{\beta}_1)\) and \(\mathrm{Beta}(\hat{\alpha}_0,\hat{\beta}_0)\), the present study evaluates the separation between meritorious and non-meritorious sentiment profiles using two complementary measures: the KL divergence \citep{KullbackLeibler1951} and the squared Hellinger distance \citep{Hellinger1909}. These metrics assess whether the full distribution of negative sentiment intensity differs across complaint outcomes, rather than relying only on differences in mean sentiment scores.

\subsubsection{Kullback--Leibler divergence}
Let \(f_1\) denote the fitted Beta density for meritorious complaints and \(f_0\) denote the fitted Beta density for non-meritorious complaints. The KL divergence from the meritorious sentiment distribution to the non-meritorious sentiment distribution is defined as
\begin{equation}
    \label{eq:kl_divergence}
    D_{\mathrm{KL}}
    \left(
    \mathrm{Beta}(\hat{\alpha}_1,\hat{\beta}_1)
    \,\|\,
    \mathrm{Beta}(\hat{\alpha}_0,\hat{\beta}_0)
    \right)
    =
    \int_{0}^{1}
    f_1(s)
    \log
    \frac{f_1(s)}{f_0(s)}
    \,ds.
\end{equation}
The KL divergence is asymmetric, so \(D_{\mathrm{KL}}(P\|Q)\) and \(D_{\mathrm{KL}}(Q\|P)\) need not be equal. In this setting, \eqref{eq:kl_divergence} measures the information loss incurred when the non-meritorious sentiment distribution is used to approximate the meritorious sentiment distribution.

For two Beta distributions, \eqref{eq:kl_divergence} has the closed form expression \citep{KBJ2000}
\begin{equation}
    \begin{split}
    D_{\mathrm{KL}}
    \left(
    \mathrm{Beta}(\hat{\alpha}_1,\hat{\beta}_1)
    \,\|\,
    \mathrm{Beta}(\hat{\alpha}_0,\hat{\beta}_0)
    \right)
    &=
    \log
    \frac{B(\hat{\alpha}_0,\hat{\beta}_0)}
         {B(\hat{\alpha}_1,\hat{\beta}_1)}
    +
    (\hat{\alpha}_1-\hat{\alpha}_0)
    \psi(\hat{\alpha}_1)\\
    &\quad
    +
    (\hat{\beta}_1-\hat{\beta}_0)
    \psi(\hat{\beta}_1)\\
    &\quad
    +
    (\hat{\alpha}_0-\hat{\alpha}_1+\hat{\beta}_0-\hat{\beta}_1)
    \psi(\hat{\alpha}_1+\hat{\beta}_1),
    \end{split}
\end{equation}
where \(B(\cdot,\cdot)\) is the Beta normalizing constant defined in \eqref{eq:beta_pdf}, and \(\psi(\cdot)=\Gamma'(\cdot)/\Gamma(\cdot)\) is the digamma function. A larger value of \(D_{\mathrm{KL}}\) indicates greater directional separation between the two fitted sentiment distributions.

\subsubsection{Squared Hellinger distance}
To complement the asymmetric KL divergence, the present study also computes the squared Hellinger distance:
\begin{equation}
    \label{eq:hellinger_distance}
    H^2
    \left(
    \mathrm{Beta}(\hat{\alpha}_1,\hat{\beta}_1),
    \mathrm{Beta}(\hat{\alpha}_0,\hat{\beta}_0)
    \right)
    =
    1
    -
    \int_{0}^{1}
    \sqrt{f_1(s)f_0(s)}
    \,ds.
\end{equation}
Unlike KL divergence, \(H^2\) is symmetric and bounded, with \(H^2 \in [0,1]\). A value of \(H^2=0\) indicates identical distributions, while a value close to 1 indicates strong distributional separation.

For two Beta distributions, \eqref{eq:hellinger_distance} reduces to the closed form expression \citep{KBJ2000}
\begin{equation}
    H^2
    \left(
    \mathrm{Beta}(\hat{\alpha}_1,\hat{\beta}_1),
    \mathrm{Beta}(\hat{\alpha}_0,\hat{\beta}_0)
    \right)
    =1-
    \frac{
    B
    \left(
    \dfrac{\hat{\alpha}_1+\hat{\alpha}_0}{2},
    \dfrac{\hat{\beta}_1+\hat{\beta}_0}{2}
    \right)
    }
    {
    \sqrt{
    B(\hat{\alpha}_1,\hat{\beta}_1)
    B(\hat{\alpha}_0,\hat{\beta}_0)
    }
    }.
\end{equation}
The squared Hellinger distance therefore provides a bounded and symmetric measure of the statistical distance between meritorious and non-meritorious sentiment distributions. Together, \(D_{\mathrm{KL}}\) and \(H^2\) form the distributional evaluation layer of the present methodology.

\subsection{Record level anomaly diagnostics}
The separation metrics compare the two outcome groups as whole distributions. The record level diagnostics developed here instead read each complaint against those fitted distributions, and they extend the logic of \citet{Gao2024}. In previous studies, a central concern is that a complaint may appear inside a meritorious classification set while still behaving like a non-meritorious complaint under the quantitative structure of the data. Anomaly detection is therefore used as a post classification reliability check, rather than as a simple search for extreme observations. The present paper adapts this idea to the distributional sentiment setting. An anomaly is defined as a complaint narrative whose observed outcome label is inconsistent with the sentiment distribution under which its corrected negative sentiment score is most plausible.

This definition links the company response outcome \(Y_i\) with the corrected sentiment score \(s_i^*\). The label \(Y_i\) records whether complaint \(i\) is meritorious, while \(s_i^*\) records the intensity of negative sentiment in its narrative. If a complaint is labeled as meritorious but its sentiment score is more consistent with the non-meritorious sentiment distribution, or if a non-meritorious complaint is more consistent with the meritorious sentiment distribution, then the complaint is treated as an outcome sentiment inconsistency. This is a record level diagnostic because each complaint is evaluated by its own tuple \((c_i,A_i,Y_i,s_i^*)\). Repeated sentiment score values do not create an indexing problem, since the complaint record \(i\) provides the one to one link between the narrative, dollar amount, outcome label, and sentiment score.

\subsubsection{Outcome sentiment inconsistency}
Let \(F_1=\mathrm{Beta}(\hat{\alpha}_1,\hat{\beta}_1)\) denote the fitted sentiment distribution for meritorious complaints, and let \(F_0=\mathrm{Beta}(\hat{\alpha}_0,\hat{\beta}_0)\) denote the fitted sentiment distribution for non-meritorious complaints. Let \(f_1\) and \(f_0\) be the corresponding probability density functions. For complaint \(i\), the density assigned by the observed outcome group is
\begin{equation*}
    f_{Y_i}(s_i^*)=
    \begin{cases}
    f_1(s_i^*), & Y_i=1,\\
    f_0(s_i^*), & Y_i=0,
    \end{cases}
\end{equation*}
and the density assigned by the opposite outcome group is
\begin{equation*}
    f_{1-Y_i}(s_i^*)=
    \begin{cases}
    f_0(s_i^*), & Y_i=1,\\
    f_1(s_i^*), & Y_i=0.
    \end{cases}
\end{equation*}
The outcome sentiment inconsistency score is defined as the log likelihood contrast
\begin{equation}
    \label{eq:likelihood_contrast}
    \ell_i=\log f_{1-Y_i}(s_i^*)-\log f_{Y_i}(s_i^*).
\end{equation}
A positive value of \(\ell_i\) indicates that the sentiment score \(s_i^*\) is more plausible under the opposite outcome distribution than under the observed outcome distribution.

For a threshold \(\eta \geq 0\), complaint \(i\) is classified as an outcome inconsistency anomaly when
\begin{equation}
    \label{eq:outcome_inconsistency}
    Z_i^{\mathrm{inc}}=\mathbf{1}\{\ell_i>\eta\}=1.
\end{equation}
The baseline specification sets \(\eta=0\), so that a complaint is flagged whenever the opposite outcome distribution assigns higher density to its sentiment score. Stricter thresholds are considered as robustness checks. The empirical anomaly rate is then
\begin{equation*}
    \widehat{Z}_{\mathrm{inc}}(\eta)=\frac{1}{N_{\mathrm{clean}}}
    \sum_{i \in \mathcal{D}_{\mathrm{clean}}} \mathbf{1}\{\ell_i>\eta\}.
\end{equation*}
This rate measures the share of complaints whose sentiment profile is distributionally inconsistent with the observed outcome label.

This rule is the direct methodological bridge from \citet{Gao2024} to the present paper. The earlier paper searches for systematic non-meritorious patterns inside a predicted meritorious set. The present paper searches for sentiment distributional inconsistency between the observed outcome label and the fitted outcome specific sentiment distribution. Thus, the anomaly is not defined by the label alone, nor by sentiment extremeness alone. It is defined by the disagreement between outcome information and distributional sentiment evidence.

\subsubsection{Tail based diagnostic anomalies}
The outcome inconsistency score is the main anomaly measure. A secondary diagnostic is used to identify complaints that are unusual within their own observed outcome group. Let
\begin{equation*}
    F_{Y_i}(s)=
    \begin{cases}
    F_1(s), & Y_i=1,\\
    F_0(s), & Y_i=0.
    \end{cases}
\end{equation*}
The two sided tail probability of \(s_i^*\) under its assigned distribution is
\begin{equation*}
    q_i=2\min\left\{F_{Y_i}(s_i^*),1-F_{Y_i}(s_i^*)\right\}.
\end{equation*}
For a threshold \(\tau \in (0,1)\), complaint \(i\) is classified as a tail based diagnostic anomaly when
\begin{equation}
    \label{tail_anomaly}
    Z_i^{\mathrm{tail}}=\mathbf{1}\{q_i \leq \tau\}=1.
\end{equation}
The baseline value is \(\tau=0.05\), with \(\tau \in \{0.01,0.05,0.10\}\) used for sensitivity analysis. This diagnostic does not replace the outcome inconsistency rule. Instead, it identifies complaints whose sentiment intensity lies in the tail of the distribution associated with their own outcome group.

\subsubsection{High sentiment no relief cases}
The third diagnostic focuses on non-meritorious complaints whose textual severity appears unusually high. Let $q_{0.95}^{(0)}=F_0^{-1}(0.95)$ denote the upper \(5\%\) cutoff of the fitted non-meritorious sentiment distribution, and let \(q_{0.75}^{A}\) denote the upper quartile of the extracted dollar amounts in the cleaned sample. A complaint is classified as a high sentiment no relief case when
\begin{equation}
\label{high_sentiment_non_relief}
Y_i=0,\qquad
s_i^*>q_{0.95}^{(0)},\qquad
A_i>q_{0.75}^{A}.
\end{equation}
This classification does not imply that the company response is incorrect. It identifies non-meritorious complaints whose narrative expresses unusually strong negative sentiment and whose dollar amount is relatively large. These cases are useful for operational risk screening because they combine textual severity with monetary context.

In summary, the anomaly diagnostics in the present paper are record level screening tools arranged from broad to targeted. The outcome inconsistency rule in \eqref{eq:outcome_inconsistency} is the primary and broadest screen. The tail based rule in \eqref{tail_anomaly} narrows this to unusual sentiment intensity within the observed outcome group, and the high sentiment no relief rule in \eqref{high_sentiment_non_relief} isolates the most actionable subset: non-meritorious complaints with both high negative sentiment and relatively large dollar amounts. This structure extends the post classification anomaly logic of \citet{Gao2024} from quantified input output systems to fitted sentiment intensity distributions without requiring a temporal window structure.

\section{Empirical distributional findings}
\label{resultsC}
The framework of Section~\ref{Method} is now applied to the cleaned CFPB sample. The findings follow the order of the methodology: negative sentiment scores and Beta fits first, then distributional separation between the two outcome groups, and finally record level anomaly diagnostics.

\subsection{Negative sentiment signals and Beta fits}
The first step is to give each cleaned narrative a single negative sentiment score. Many CFPB narratives run past the input limit of 512 tokens used by common financial models in the BERT family \citep{Devlin2019,Araci2019}. Cutting the text at that limit can drop the later parts of a complaint. To lower this risk, the present paper computes the score with two strategies for handling long text.
 
The first strategy splits each narrative \(c_i\) into \(K_i\) consecutive chunks of at most 512 tokens. The classifier is applied to each chunk, and the chunk score is the average
\begin{equation*}
    s_i^{\mathrm{chunk}} =\frac{1}{K_i}\sum_{k=1}^{K_i}p_{ik,\mathrm{neg}},
\end{equation*}
where \(p_{ik,\mathrm{neg}}\) is the negative sentiment probability for chunk \(k\) of complaint \(i\). The second strategy follows the compression approach of \citet{GYSZ2026}. Each narrative is first compressed with a T5 transformer \citep{Raffel2020}, and the negative sentiment probability is read from the compressed text. This gives a second score, \(s_i^{\mathrm{T5}}\).
 
The two strategies produce very similar scores. To improve stability and lean less on any single rule for long text, the present paper takes the average of the two as the final score for each document:
\begin{equation*}
    s_i=\frac{s_i^{\mathrm{chunk}}+s_i^{\mathrm{T5}}}{2}.
\end{equation*}
This score stays in \([0,1]\) and feeds the boundary correction and Beta fitting in Section~\ref{Method}.
 
The diagnostics in this section are computed on the cleaned sample and reported by outcome group, with group sizes \(N_1=85{,}501\) and \(N_0=95{,}815\) from Section~\ref{Data}.

\begin{figure}[htbp]
\centering
\subfigure[Histogram and smoothed curves for sentiment score.\label{FinBert1}]{%
\includegraphics[width=0.48\textwidth]{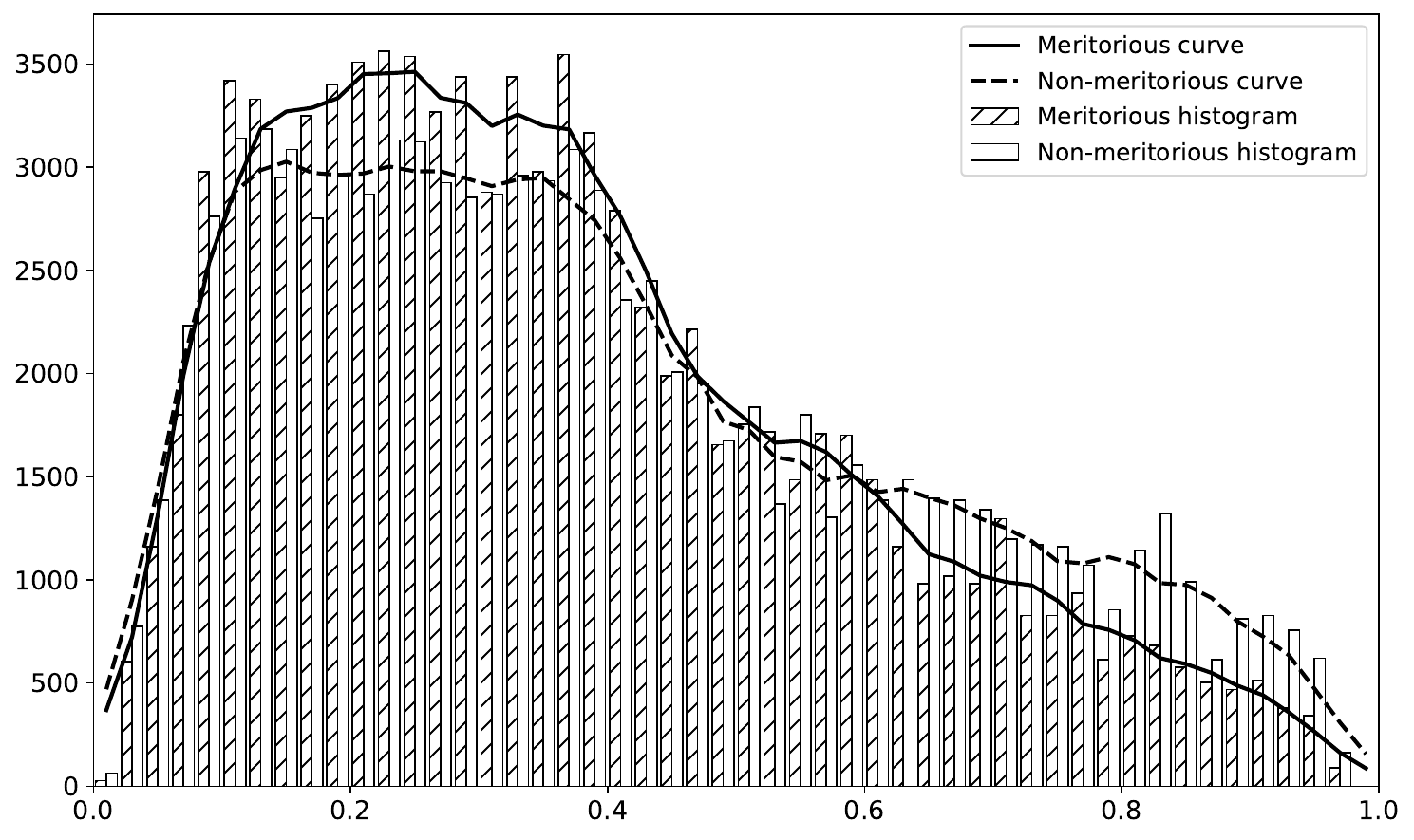}}%
\hfill
\subfigure[Beta distribution fits for sentiment score.\label{FinBert2}]{%
\includegraphics[width=0.48\textwidth]{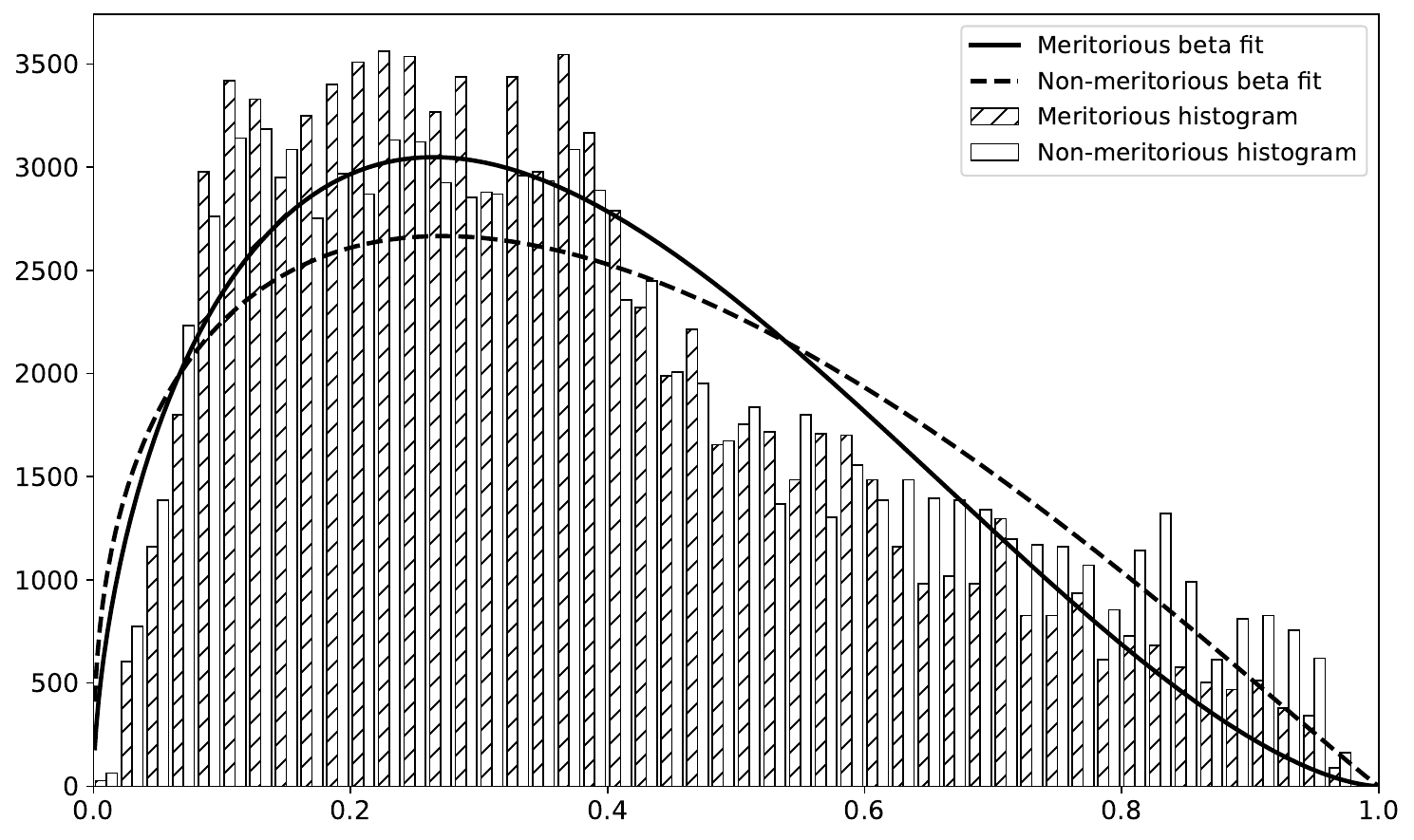}}%
\caption{Negative sentiment score distributions and fitted Beta distributions by complaint outcome.}
\label{NegSentiment}
\end{figure}

Figure~\ref{NegSentiment}\textup{(a)} 
plots the averaged scores by outcome. The meritorious and non-meritorious distributions have similar shapes. Most scores sit in the lower to middle range, roughly \(0.10\) to \(0.50\), and both groups have a right tail reaching toward \(1\). The non-meritorious curve is a little heavier at the high end, above about \(0.60\), while the meritorious curve is more concentrated in the lower to middle range.

The averaged score \(s_i\) lies in \([0,1]\) and the corrected score \(s_i^*\) lies in \((0,1)\), so the scores meet the support that the Beta distribution needs. The corrected scores are fitted for the full cleaned sample and separately for the two outcome groups. Table~\ref{beta_fit_summary}
\begin{table}[htbp]
\centering
\caption{Beta fits for negative sentiment scores in the cleaned sample and by outcome group.}
\label{beta_fit_summary}

\footnotesize
\setlength{\tabcolsep}{5pt}
\renewcommand{\arraystretch}{1.15}

\begin{tabular}{lrrrrrrr}
\hline
Sample & \(N\) & \(\hat{\alpha}\) & \(\hat{\beta}\) & Fitted mean & Fitted var. & Sample mean & Sample var. \\
\hline
Full sample & \(181{,}316\) & \(1.473\) & \(2.296\) & \(0.391\) & \(0.050\) & \(0.382\) & \(0.052\) \\
Meritorious & \(85{,}501\) & \(1.593\) & \(2.648\) & \(0.376\) & \(0.045\) & \(0.367\) & \(0.046\) \\
Non-meritorious & \(95{,}815\) & \(1.394\) & \(2.058\) & \(0.404\) & \(0.054\) & \(0.395\) & \(0.057\) \\
\hline
\end{tabular}
\end{table}
reports the fitted parameters together with the fitted and sample moments, with group sizes \(N_1=85{,}501\) and \(N_0=95{,}815\) from Section~\ref{Data}.
 
The full sample fit gives the overall profile of negative sentiment in the cleaned corpus. The full sample mean sits between the two groups and closer to the larger non-meritorious group. The group specific fits then show how this profile differs by outcome: the non-meritorious group has a slightly higher fitted mean and a larger fitted variance, while the two group distributions remain close. The small gaps between the fitted and sample moments mean that the fitted Beta distributions are compact summaries, not exact matches to the sample moments.

Figure~\ref{NegSentiment}\textup{(b)} lays the fitted Beta curves over the histograms. The two curves are close, but the non-meritorious one has a slightly higher mean and a larger variance. Thus, non-meritorious complaints tend to have somewhat stronger and more spread out negative sentiment.

\begin{figure}[htbp]
\centering
\includegraphics[width=0.92\textwidth]{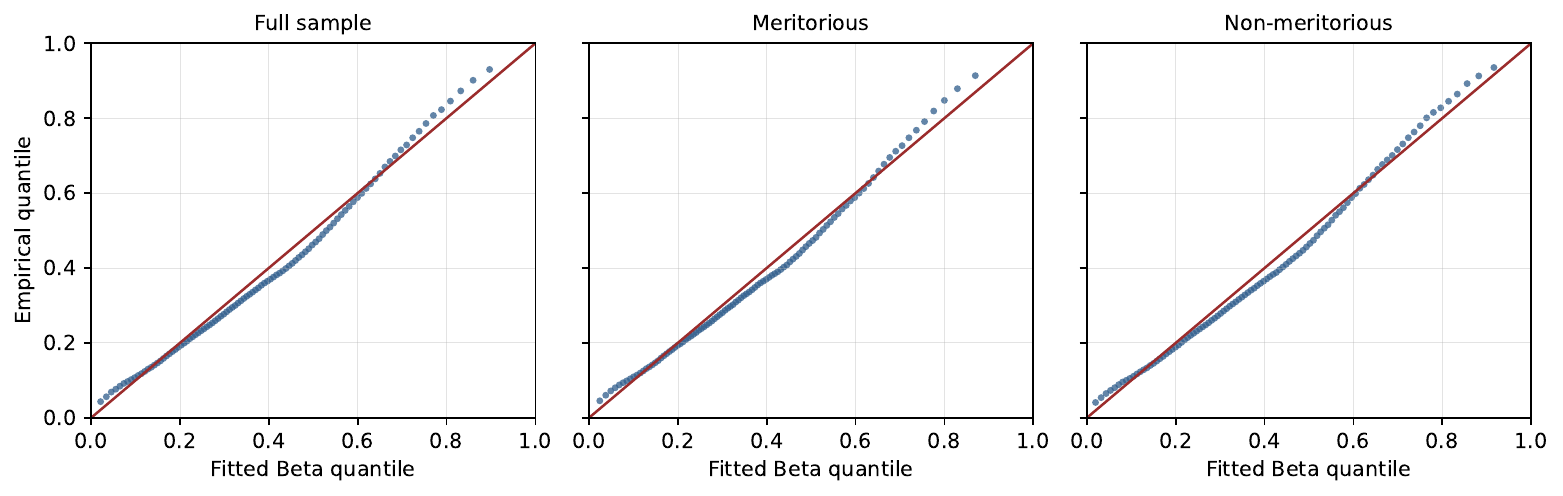}
\caption{QQ diagnostics for the fitted Beta models of negative sentiment scores. Each panel plots empirical against fitted Beta quantiles for the full sample and the two outcome groups; the red line is the \(45^\circ\) reference.}
\label{QQplot}
\end{figure}

To check whether a single Beta distribution is adequate, Figure~\ref{QQplot} 
shows QQ diagnostics for the full sample and the two outcome groups. In each panel the points track the \(45^\circ\) line through the lower and central range and rise above it in the upper tail, so the empirical upper tail is a little heavier than the fitted Beta. The Kolmogorov--Smirnov statistic, the largest gap between the empirical and fitted cumulative distribution functions, is \(0.058\) for the full sample, \(0.056\) for the meritorious group, and \(0.055\) for the non-meritorious group. With samples this large, a formal test rejects the exact Beta null for any small deviation, so these values are read as effect sizes rather than as test decisions: the misfit is small in the center and concentrated in the upper tail. The single Beta model therefore gives a useful first approximation of the bounded sentiment distribution, while the upper tail structure it leaves is exactly what the tail and high sentiment diagnostics in Section~\ref{AnomalyD} examine.
 
The gap between the two groups is too small to separate them on its own, which is why the present paper uses distributional measures rather than visual inspection or mean comparison. These measures are reported next.

\subsection{Distributional separation between outcome groups}
The fitted Beta distributions are compared with the KL divergence and the squared Hellinger distance from Section~\ref{Method}. Table~\ref{DistributionSSS}
\begin{table}[htbp]
\centering
\caption{Distributional separation between the fitted Beta sentiment distributions. \(F_1\) and \(F_0\) denote the meritorious and non-meritorious fits.}
\label{DistributionSSS}
\begin{tabular}{lr}
\hline
Metric & Value \\
\hline
\(D_{\mathrm{KL}}(F_1 \,\|\, F_0)\) & \(0.0168\) \\
\(D_{\mathrm{KL}}(F_0 \,\|\, F_1)\) & \(0.0202\) \\
Symmetric KL average & \(0.0185\) \\
Squared Hellinger distance (\(H^2\)) & \(0.0046\) \\
\hline
\end{tabular}
\end{table}
reports the values. The KL divergence from the meritorious to the non-meritorious distribution is \(0.0168\), and the reverse is \(0.0202\). The reverse is a little larger, which fits the larger variance and heavier upper tail of the non-meritorious distribution. The symmetric average is \(0.0185\), so the two fitted distributions are close but not identical.

The squared Hellinger distance is \(0.0046\), near zero on its \([0,1]\) scale, which confirms heavy overlap. Thus, the negative sentiment score is not a separator of complaint outcomes on its own. It is a weak but measurable distributional signal, to be combined with the dollar amount and other complaint attributes in the wider framework.

\begin{figure}[htbp]
\centering
\includegraphics[width=0.75\textwidth]{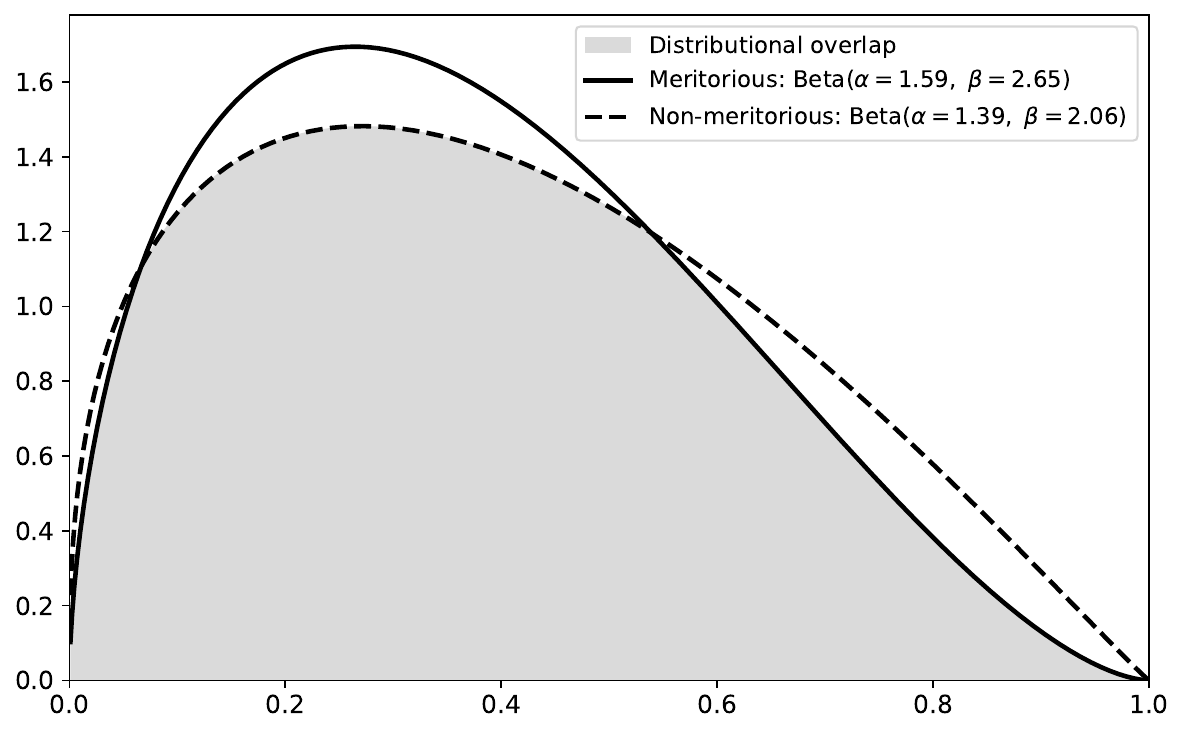}
\caption{Overlap between the fitted Beta densities for meritorious and non-meritorious complaints.}
\label{betaDensity}
\end{figure}

Figure~\ref{betaDensity}
shows this directly. The two fitted curves share a mode in the lower to middle range, and the shaded region marks the mass they share. The meritorious density is a little more peaked, while the non-meritorious density is flatter and stays higher over much of the upper range. This matches the fitted means, variances, and divergences: the non-meritorious group is slightly more dispersed, yet the two groups stay close.

\subsection{Anomaly diagnostics from fitted sentiment distributions}
\label{AnomalyD}
Because the two distances are small, the fitted distributions work better as baselines for spotting outcome sentiment inconsistency. Each complaint is indexed by its record identifier \(i\), with the tuple \((c_i,A_i,Y_i,s_i^*)\). The score \(\ell_i\) in \eqref{eq:likelihood_contrast} is computed for each record on its own and read against that record's label \(Y_i\); the dollar amount \(A_i\) enters only the high sentiment no relief diagnostic. Repeated score values cause no problem.

With the fitted parameters above, the two Beta densities cross at about $s^* = 0.0666$ and $s^* = 0.5392.$
Between these points the meritorious density is higher; outside them the non-meritorious density is higher. This follows from the larger variance of the non-meritorious distribution, which puts more mass in both tails, above all in the upper tail.

Table~\ref{AnomlayDRegion}
\begin{table}[htbp]
\centering
\caption{Anomaly regions implied by the fitted Beta distributions}
\label{AnomlayDRegion}
\begin{tabular}{lll}
\hline
Observed outcome & Sentiment region & Diagnostic interpretation \\
\hline
Meritorious 
& \(s_i^* < 0.0666\) or \(s_i^* > 0.5392\) 
& More consistent with the non-meritorious density \\

Non-meritorious 
& \(0.0666 < s_i^* < 0.5392\) 
& More consistent with the meritorious density \\
\hline
\end{tabular}
\end{table}
shows the regions for the baseline threshold \(\eta=0\). The rule is distributional by design. A meritorious complaint is flagged when its score falls where the non-meritorious density is higher, and a non-meritorious complaint is flagged when its score falls in the central band where the meritorious density is higher. Since the two distributions overlap so much, the baseline rule flags a wide set of weak inconsistencies. The present paper therefore reads \(\eta=0\) as a screening threshold and reports stricter \(\eta\) and a range of \(\tau\) as sensitivity checks.

Table~\ref{Anomaly_diagnostic_summary}
\begin{table}[htbp]
\centering
\caption{Anomaly diagnostics on the cleaned sample: baseline and threshold sensitivity.}
\label{Anomaly_diagnostic_summary}
\footnotesize
\setlength{\tabcolsep}{4pt}
\begin{tabular}{llrrrr}
\hline
Diagnostic & Threshold & Meritorious & Non-meritorious & Total & Total rate \\
\hline
Outcome inconsistency & \(\eta=0\)    & \(20{,}572\) & \(66{,}754\) & \(87{,}326\) & \(48.16\%\) \\
Outcome inconsistency & \(\eta=0.01\) & \(19{,}546\) & \(65{,}422\) & \(84{,}968\) & \(46.86\%\) \\
Outcome inconsistency & \(\eta=0.05\) & \(15{,}518\) & \(58{,}179\) & \(73{,}697\) & \(40.65\%\) \\
Outcome inconsistency & \(\eta=0.10\) & \(12{,}081\) & \(39{,}802\) & \(51{,}883\) & \(28.61\%\) \\
Tail anomaly          & \(\tau=0.01\) & \(1{,}257\)  & \(795\)      & \(2{,}052\)  & \(1.13\%\)  \\
Tail anomaly          & \(\tau=0.05\) & \(4{,}412\)  & \(4{,}561\)  & \(8{,}973\)  & \(4.95\%\)  \\
Tail anomaly          & \(\tau=0.10\) & \(7{,}977\)  & \(9{,}447\)  & \(17{,}424\) & \(9.61\%\)  \\
\hline
\end{tabular}
\end{table}
reports the baseline counts together with the threshold sensitivity. At the baseline \(\eta=0\), the outcome inconsistency rule flags about \(87{,}326\) complaints, or \(48.16\%\) of the cleaned sample. This large share is expected because the two fitted distributions overlap so much, so it is a broad screening result, not evidence of outcome error. The rule is also asymmetric across groups: it flags \(69.67\%\) of non-meritorious complaints but only \(24.06\%\) of meritorious ones, because most non-meritorious scores fall in the central band where the meritorious density is higher. This asymmetry reinforces the reading of \(\eta=0\) as a broad screen rather than a precise error detector. Raising \(\eta\) makes the rule more conservative, and the flagged total falls to \(51{,}883\), or \(28.61\%\), at \(\eta=0.10\). The tail rule is more selective: at the baseline \(\tau=0.05\) it flags about \(8{,}973\) complaints, or \(4.95\%\), and its total rises from \(1.13\%\) at \(\tau=0.01\) to \(9.61\%\) at \(\tau=0.10\). Because each tail rate sits close to its \(\tau\), the fitted distributions give reasonable tail calibration for this diagnostic. These patterns support \(\eta=0\) and \(\tau=0.05\) as baseline screening choices rather than as the only specification.

The most actionable class is the complaint with high negative sentiment and no relief. These have \(Y_i=0\), a score in the upper tail of the non-meritorious distribution, and a dollar amount \(A_i\) that is large for the cleaned sample. Such complaints are not necessarily mislabeled. The narrative carries unusually strong negative sentiment and a sizable amount, yet the company response is closed with an explanation. This group matters for operational risk monitoring, since it points to complaints that read as severe even though no monetary or non-monetary relief is recorded.
This class is defined by the upper \(5\%\) cutoff of the fitted non-meritorious distribution and the upper quartile of the dollar amount, that is \(s_i^*>0.814\) and \(A_i>2{,}861.54\). This gives \(1{,}629\) complaints, about \(1.70\%\) of the non-meritorious group and \(0.90\%\) of the cleaned sample.

Overall, our anomaly analysis therefore provides a direct diagnostic at the record level, which identifies complaints whose sentiment differs from the distribution associated with their fitted outcome. When combined with dollar amounts and other attributes, the resulting score provides a practical screening tool for identifying records that require further review.

\section{Conclusion}
\label{conclusioned}
The present paper builds a distributional sentiment framework for CFPB consumer complaint narratives. The sample covers complaints filed between 1 September 2023 and 30 June 2026. It is drawn from the ``Credit reporting or other personal consumer reports'' category and keeps narratives that report a dollar amount between 1 and 10{,}000 dollars. Each narrative is turned into a continuous negative sentiment score by a transformer classifier that handles long text. The scores are treated as bounded random variables and modeled with Beta distributions, both for the full sample and for the meritorious and non-meritorious groups. The two fitted group distributions are empirically close, as shown by the small KL divergences and squared Hellinger distance. The non-meritorious group has a slightly higher fitted mean, a larger variance, and a heavier upper tail. Because the two distributions overlap substantially, a comparison of the group distributions cannot, by itself, separate meritorious from non-meritorious complaints. Anomaly diagnostics are therefore best positioned to complement operating at the case level to identify complaints that group wise comparison necessarily overlooks.

The main contribution is to move SA in complaint research from discrete polarity labels toward continuous distributional measurement. The present paper does not use sentiment only as an isolated covariate or a classification input. It models the full shape of negative sentiment and compares the two outcome groups through probability distributions, divergence measures, and density based anomaly diagnostics. The group comparison shows that the two fitted distributions are close, so the sentiment signal is not a sharp group separation rule. The anomaly layer therefore provides the main diagnostic use of the framework at the level of the single complaint record. It flags complaints whose observed outcome does not match the fitted sentiment distribution, especially complaints with high negative sentiment that received no relief. Such cases may deserve closer review once the dollar amount and other attributes are considered. This structure ties together three parts that are usually studied separately: sentiment extraction from text, bounded response modeling, and operational risk monitoring. The present paper therefore gives a clear statistical way to use complaint narratives as evidence in consumer finance and operational risk work.

\subsection{Limitations}
The present paper covers one main CFPB product category and one sample period that starts after August 2023. This choice improves consistency because the analysis is conducted after the CFPB complaint form update and after the main COVID-19 emergency period. The findings should still be read within this setting. Credit reporting and personal consumer report complaints make up a large share of recent CFPB records, yet other categories may differ in narrative style, company response patterns, dollar amount distributions, and sentiment structure.

The sentiment score also depends on the model. A transformer classifier is a practical way to turn long narratives into continuous negative sentiment probabilities, but the scores can still reflect model calibration, domain transfer, text compression, and the way chunks are combined. The Beta distribution gives an interpretable model for bounded scores, yet a fitted distribution is a summary of complex narrative evidence, not a full account of what a complaint means. For this reason, the anomaly diagnostics are read as screening signals. They are not proof that a company response is wrong.

\subsection{Future work}
Future work can extend the framework to more CFPB product categories, longer periods, and finer subgroups, such as company, issue type, state, submission channel, and consumer demographics when available. This would show whether the gap between meritorious and non-meritorious complaints stays stable across settings. When complete ordered time windows are available, future studies can also compare monthly or quarterly fitted distributions with a reference distribution through the KL divergence, the squared Hellinger distance, or an index based distributional shift rule \citep{GribkovaZitikis2018}.

A second direction is to join sentiment distribution modeling with severity and outcome models. The present paper treats negative sentiment and dollar amount as related but separate signals. Future studies can test whether the joint behavior of \(s_i^*\), \(A_i\), complaint attributes, and company response improves the prediction of relief outcomes, loss severity, or operational risk escalation. Human review of complaints with high negative sentiment and no relief would also help. It would show whether the anomaly diagnostics point to complaints that matter, not just statistical tail cases.

A third direction is to compare other sentiment scoring models and other distributional families. The present framework uses negative sentiment probabilities from a transformer and Beta distributions because the scores are bounded on \((0,1)\). Future research can test whether language models adapted to the domain, calibrated probability scores, mixture Beta models, Beta models inflated at zero and one, or semiparametric density estimators fit complaint narratives better. Such a comparison would show whether the overlap between the meritorious and non-meritorious distributions comes from the complaint process itself or from the modeling choices used to summarize tone.


\end{document}